# RRT-Rope: A deterministic shortening approach for fast near-optimal path planning in large-scale uncluttered 3D environments

Louis Petit[1] and Alexis Lussier Desbiens[1]

***Abstract*— Many path planning algorithms have been introduced so far, but most are costly, in path cost and in processing time, in large-scale uncluttered 3D environments such as underground mining stopes explored by an unmanned aerial vehicle (UAV). Rapidly-exploring Random Tree (RRT) algorithms are popular because of their probabilistic completeness and rapidity in finding a feasible path in single-query problems. Many of the algorithms (e.g. Informed RRT*, RRT#) developed to improve RRT need considerable time to converge in large environments. Shortcutting an RRT is an old idea that has been proven to outperform RRT variants. This paper introduces a new method, RRT-Rope, that aims at finding a near-optimal solution in a drastically shorter amount of time. The proposed approach benefits from fast computation of a feasible path with an altered version of RRT-connect, and post-processes it quickly with a deterministic shortcutting technique, taking advantage of intermediate nodes added to each branch of the tree. This paper presents simulations and statistics carried out to show the efficiency of RRT-Rope, which gives better results in terms of path cost and computation time than other popular RRT variations and shortening techniques in all our simulation environments, and is up to 70% faster than the next best algorithm in a representative stope.**

## I. INTRODUCTION

The challenges of 3D path planning have been addressed by many [1], with Rapidly-exploring Random Tree (RRT) algorithms being commonly used. However, although RRTs require low computation time to find a first collision-free trajectory, the time required to find an optimal (Euclidean shortest) path can be very large, especially in large-scale uncluttered 3D environments such as underground mining stopes. The problem of navigating through stopes can be considered holonomic for a mapping unmanned aerial vehicle (UAV), as described in section II-A. Node based algorithms like A* [2] are frequently used for 3D holonomic path planning but the computational complexity makes them slow in large environments for small discretization of the 3D space, as shown in Section V-A, while large discretization makes them incomplete or non-optimal.

Traditional RRTs [3] are based on generating random points in the reachable space. A connection is then established from the nearest node in the tree $x_{near}$ to a point $x_{new}$, which is located on the line between $x_{near}$ and a point of randomly generated coordinates $x_{rand}$, at an infinitesimal step $\epsilon$ from $x_{near}$. One of the main reasons why RRTs are so widely used is that they are biased towards search into the largest Voronoi regions in a graph of the configuration space. In holonomic problems, the infinitesimal step size ($\epsilon$) can be omitted, and $x_{rand}$ used as $x_{new}$, due to the high probability that the generated random point is reachable from the nearest node [4]. RRT algorithms are popular because of their probabilistic completeness and quick path computation in single-query problems, but they fail to converge to optimal solutions because of existing graph bias.

RRT* is an evolution of RRT that ensures asymptotic optimality [5]. It adds recursive rewiring to RRT by connecting new nodes as parent nodes of a neighbour node if the latter's cost is improved. Informed RRT* [6] improves the convergence rate and solution quality by generating new random points in an ellipsoidal subset of the space. Other ideas for online path optimization, including RRT# [7], C-FOREST [8], searching around beacons [9], and graph pruning [10] [11] have been presented. First, it is shown in Section V-A that the final path cost of RRT-Rope does not depend on initial path quality and online optimization is therefore obsolete. Second, if the goal position is challenging to reach, all these algorithms need large computation time to find a first path. To that end, RRT-connect [12] was developed to improve path-finding time by growing two trees simultaneously toward each other: $T_A$, a tree from $x_{start}$; and $T_B$, a tree from $x_{goal}$. This algorithm is efficient for fast single-query path finding, but not optimal.

Post-processing techniques have been introduced to overcome this problem. Shortcutting an RRT is an old idea that has been proven to outperform optimal RRT variants [13]. Node pruning, based on triangular inequality, has been used on A* trajectories [9], but A* inherent computational complexity is an apparent disadvantage in high dimensions. Node pruning is also used with RRT* [14]. However, the resulting path is not optimal because the node resolution is reduced through node pruning.

Partial-shortcut [15] is one of the most popular shortening methods. It selects 2 arbitrary points in a path and moves the path segment between them on a straight line (or linear interpolation for multiple DOF systems) if there is no collision. Since points are moved instead of pruned, the path is closer to optimality. However, computation time is not optimal due to some irrelevant shortcuts performed because of the non-deterministic point selection (e.g. for an already straight line path or for a to-be pruned portion of path).

An iterative technique based on bisection was developed in [16] but it could not find a better solution located far away and will only return convex solutions.

Elastic strips [17] is another iterative technique that operates by locally displacing a point according to a virtual

This work was supported by the University of Sherbrooke.

[1]The authors are with the Createk Design Lab, University of Sherbrooke, Sherbrooke, J1K 2R1, Canada louis.petit@usherbrooke.ca; alexis.lussier.desbiens@usherbrooke.ca

attractive force towards its two neighbours and a repulsive force coming from obstacles. Since the algorithm is designed for online control, the convergence is slow when used as a shortening method, even with inertial gradient descent.

Finding a feasible path is thus generally very efficient with RRT-like techniques like RRT-connect, but optimization remains lengthy in high-dimensional problems. In large configuration spaces, most shortening methods require many iterations. Moreover, the algorithm typically ends when the computation time exceeds a value fixed by the user, or when the measured convergence rate is small enough. This makes for variable and unpredictable performance.

A new method, *RRT-Rope*, is introduced in this paper to rapidly optimize the first path found with an RRT-connect variant and a deterministic node selection.

This paper is structured as follows. Section 2 establishes the notation and formal definition of the problem. Section 3 details the operation of the RRT-Rope algorithm. In Section 4, the algorithm's complexity and optimality are analyzed. Section 5 presents the results of our simulations with a mapping UAV in large stopes. A comparison with other methods is also presented—with a focus on convergence, performance relative to the other shortening methods cited in this section, and step-size sensitivity—to highlight the RRT-Rope algorithm's efficiency in stopes. Finally, this paper concludes with potential extensions to the RRT-Rope algorithm.

## II. PRELIMINARY MATERIAL

### A. Assumptions

Since the UAV is used for mine mapping, the problem of path planning in mine stopes relies on the following assumptions. For a UAV, the Euclidean space is $SE(3) = \mathbb{R}^3 \times SO(3)$ for a 6-parameter configuration: three positions $(x, y, z)$ and three angles $(\phi, \theta, \psi)$. Since a mapping drone would likely travel at low velocities, the roll $(\phi)$ and pitch $(\theta)$ angles are close to zero. Furthermore, at slow speeds, the yaw angle $(\psi)$ can be controlled independently to, e.g., always let the sensors point forward. Finally, the obstacles in the environment are enlarged offline by the drone radius and a safety factor. This allows us to consider the drone as a single point. For these reasons, the search space can be reduced to $\mathbb{R}^3$ for a point-like problem and dynamic constraints can be ignored. Therefore, RRT can be used without infinitesimal step size $(\epsilon)$ as mentioned in [4].

### B. Notation

Following the notation introduced in [18], we can define the world $\mathcal{W}$, a semi-algebraic obstacle region $\mathcal{O} \in \mathcal{W}$, a semi-algebraic UAV $\mathcal{A}$ defined in $\mathcal{W}$ as a point, and $q$ the configuration of $\mathcal{A}$. In the case of a drone exploring stopes, $\mathcal{W} = \mathbb{R}^3$ and $q = (x_t, y_t, z_t)$.

The configuration space $C$ can be decomposed into the free space $C_{free}$ and the obstacle region $C_{obs}$. Practically, $C_{free}$ and $C_{obs}$ are obtained with the voxel occupancy probability in an OctoMap [19] generated from *LiDAR* measurements enlarged by the drone radius. The obstacle region $C_{obs} \subseteq C$ is defined as $C_{obs} = \{q \in C \mid \mathcal{A}(q) \cap \mathcal{O} \neq \emptyset\}$. The resulting set of permissible sets is $C_{free} = C \setminus C_{obs}$. Since $C_{free}$ is an open set and $C_{obs}$ is closed, if $\mathcal{A}$ touches $\mathcal{O}$ only by a boundary intersection,

$$int(\mathcal{A}(q)) \cap int(\mathcal{O}) = \emptyset \text{ and } \mathcal{A}(q) \cap \mathcal{O} \neq \emptyset. \quad (1)$$

The query pair $(q_I, q_G)$ includes the initial configuration $q_I \in C_{free}$ and the final configuration $q_G \in C_{free}$.

### C. Problem formulation

For feasible path planning, a complete algorithm must compute a continuous path, $\tau : [0, 1] \to C_{free}$, such that $\tau(0) = q_I$ and $\tau(1) = q_G$ or correctly report that such a path does not exist. In our case, a path always exists thanks to the stope geometry pictured in Fig. 1.

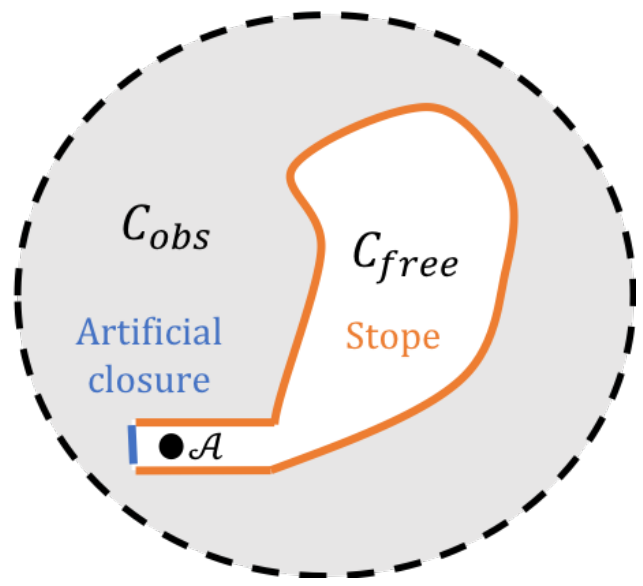

Fig. 1: Path-planning problem in a mine stope.

The topological space $C_{free}$ is bounded, connected (i.e., it cannot be represented as the union of two or more disjoint non-empty open sets) and path-connected, since for all $q_I, q_G \in C_{free}$, there exists a path $\tau$ such that $\tau(0) = q_I$ and $\tau(1) = q_G$. In stopes, there is no column running from one end to the other. Then, $C_{free}$ is simply connected (i.e., any loop can be reduced to a point without encompassing any domain not included in the space). For the definition to ban rocks in suspension, it will be said that the stope surface is connected, and the stope volume is simply connected. Finally, the stope geometry can be considered as mainly wide, from 3 to 30 m wide and 30 to 200 m deep. As [20] stated, shortcutting is unlikely to discover a new homotopy. Considering that mine stopes are 3-ball homotopic environments, all paths are homotopic and thereby any path is in the optimal homotopy class.

## III. RRT ROPE

For problems where $\tau$ can be a straight line (trivial paths), optimality is met, and the path cost is the Euclidean distance in $\mathbb{R}^3$. In any other case, if $\tau$ is optimal, there is at least one $\tau(i) = q_i$ with $i \in [0, 1]$, such that (1) is verified. The path must then be defined on the closure of $C_{free}$.

In a simple environment, an RRT without infinitesimal step size $(\epsilon)$ could find the yellow path in Fig. 2a. Meanwhile, the blue path would be the optimal path for the point-like problem without dynamic constraints. This path is composed of straight lines and parts of the closure of $C_{free}$. Since RRT-connect has been found to be fast for single-query path planning, one could optimize the first feasible path found by finding the mapping $\tau_f \to \tau^*$. RRT-Rope achieves this in a way that is similar to applying tension to a rope.

### A. Path finding

The first step of RRT-Rope is similar to RRT-connect [12] with the exception that it omits $\epsilon$ and inserts intermediate nodes to each branch, during the path search, with a step size $\delta$ between them as shown in Fig. 2b. Since a non-collision ray casting check has first been performed to wire a branch, it is obvious that all points of a tree branch are in $C_{free}$. For holonomic path planning in large environments, this is faster than RRT with $\epsilon$ (assumed similar to $\delta$) that would require more iterations of random node generation.

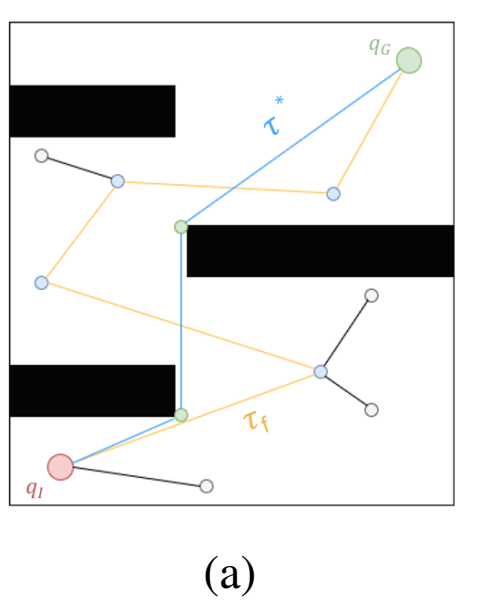

(a)

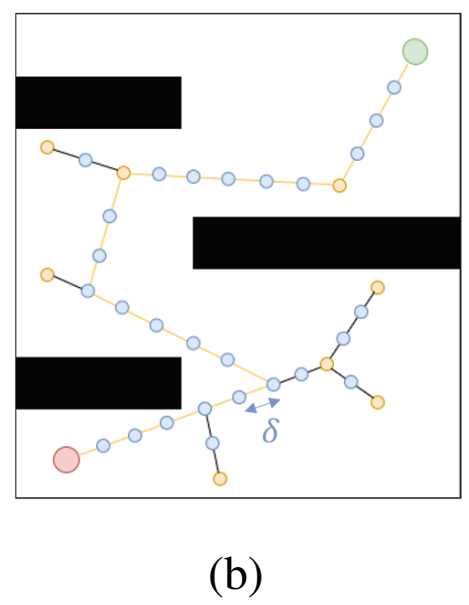

(b)

Fig. 2: Path planning in a simple environment. Fig. (a) shows the first path found by RRT (yellow) and the optimal path (blue) from $q_I$ (red) to $q_G$ (green). Fig. (b) shows the same feasible path with intermediate nodes (blue).

### B. Path optimization

The idea of the rewiring algorithm (depicted in Fig. 3) benefits from intermediate nodes to try to get as close as possible to the closure of $C_{free}$. This is achieved through a method reminiscent of the tightening of a rope. Since the problem is holonomic, the optimal path will be the special case of a Dubins path that does not impose any minimum radius for the robot. In these conditions, the minimal radius of the rope turns is imposed by the obstacles if they are smooth, or is zero otherwise.

For each path node $i$, taken one by one from $q_I$ to $q_G$, a collision check is performed with each node $j$ starting from $q_G$ to $i$. It takes advantage of the width of the environment by checking the farthest nodes first, such as a rope whose curvature would be constrained by obstacles, which ensures to only perform relevant shortcuts, unlike random selection. If the shortcut is collision-free, intermediate nodes will be inserted and the $(i, j)$ pair will be connected by a straight line. Thus, node resolution is ensured to be constant along the path unlike some other algorithms. For the sake of efficient computation, when the path is already a straight-line between $i$ and $j$, this will be bypassed to enhance memory management. In both cases, the next collision check will start at $i + 1$. This is performed by Algorithm 2.

Algorithm 1 shows the full RRT-Rope procedure. It operates directly on the trees instead of the path, since the nodes could be used for subsequent RRT rewiring (such as Informed RRT*), as discussed later in section VI, but one could use a simplified version on a given path. For the sake of memory management, $\tau_{raw}$ only stores indexes of nodes and whether it belongs to $T_A$ or $T_B$.

There are three possible shortcuts with RRT-connect: inside $T_A$, inside $T_B$, or between $T_A$ and $T_B$. Lines 10-12 handle the first case, whereas lines 14-15 handle the two other cases. If $\tau_{raw}[i] \in T_A$ and $\tau_{raw}[j] \in T_B$, the intermediate nodes of the new branch will be added to $T_B$ and the connection node will change accordingly to become $q_i$.

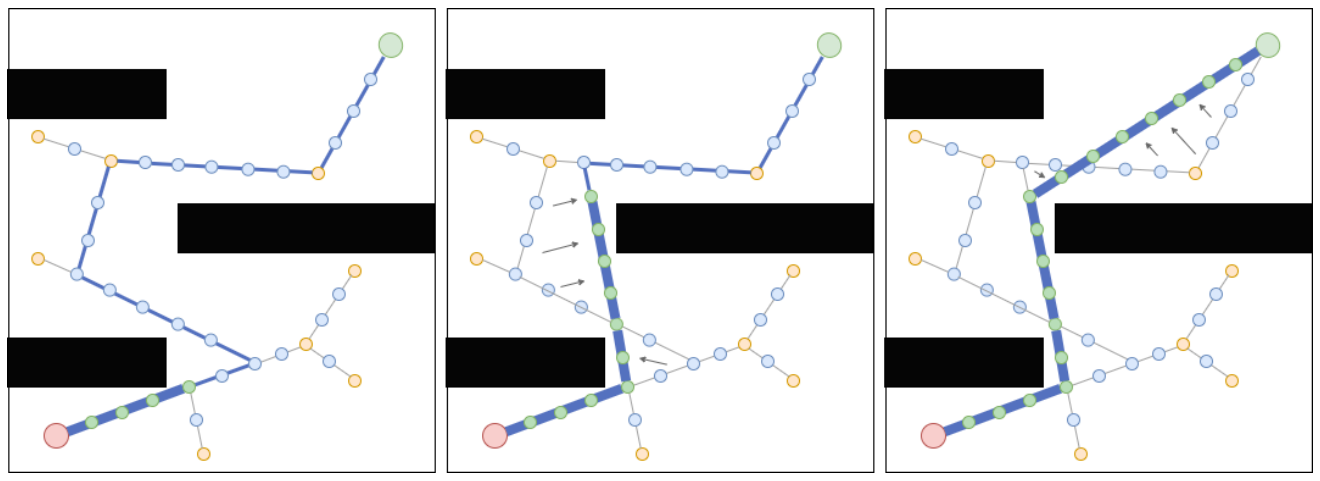

Fig. 3: Sequence of shortening algorithm, from left to right.

**Algorithm 1** RRT-Rope($\delta$)

1: $(T_A, T_B) \leftarrow$ RRT-connect($\delta$, $\epsilon$ = false);
2: $\tau_{raw}$.init(treeToPath($q_I, T_A$));
3: $\tau_{raw}$.add(treeToPath($q_{connect}, T_B$));
4: **for** $i \leftarrow 0$ **to** length($\tau_{raw}$) **do**
5:     $\tau_{opti}$.add($\tau_{raw}[j]$);
6:     **for** $j \leftarrow$length($\tau_{raw}$) **to** $i + 1$ **do**
7:         **if** collisionFree($q_i, q_j$) **then**
8:             **if** betterPathCost($q_i, q_j$) **then**
9:                 **if** $\tau_{raw}[i] \in T_A$ **and** $\tau_{raw}[j] \in T_A$ **then**
10:                     $T_A$.insertIntermediateNodes($\delta, q_i, q_j$);
11:                     $\tau_{raw}$.init(treeToPath($q_i, T_A$);
12:                     $\tau_{raw}$.add(treeToPath($q_{connect}, T_B$));
13:                 **else**
14:                     $T_B$.insertIntermediateNodes($\delta, q_i, q_j$);
15:                     $\tau_{raw}$.init(treeToPath($q_i, T_B$);
16:                 $i \leftarrow -1$;
17:             **break**;
18: **return** $\tau_{opti}$;

**Algorithm 2** betterPathCost($q_i, q_j$)

1: **if** $(q_i, q_j) \in T_A$ **or** $(q_i, q_j) \in T_B$ **then**
2:     **return** dist($q_i, q_j$) + $q_i$.cost $<$ $q_j$.cost;
3: **else if** $q_i \in T_A$ **and** $q_j \in T_B$ **then**
4:     **return** dist($q_i, q_j$) + $q_i$.cost$_A$ + $q_j$.cost$_B$ $<$ $\tau_{raw}$.cost;

Fig. 4 shows a simulation performed in a 3D stope. One can see that only three branches (two green in tree A, and one pink in tree B) were needed to find a first path. The light green branches come from RRT-Rope's back-to-back shortcuts.

## IV. ANALYSIS

This section presents an analysis of the algorithm. The complexity of adding intermediate nodes and the optimization step are evaluated in the first sub-section, then completeness and optimality are discussed.

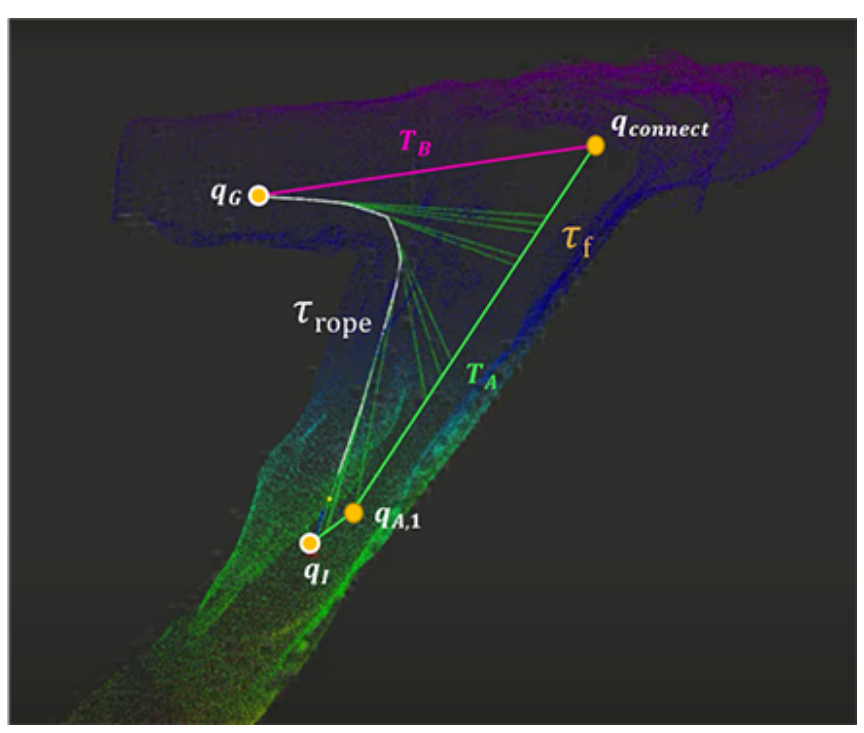

Fig. 4: Resulting path (white) of the RRT-Rope algorithm in a stope, after optimizing tree A (green) and tree B (pink).

### *A. Complexity*

The algorithm's complexity is composed of two parts: search and optimization.

*1) Path finding:* As stated in [21] for RRT and RRT*, the space complexity is $\mathcal{O}(n)$ and the time complexity is $\mathcal{O}(n\,log(n))$, expressed as a number of samples $n$.

Here, $n$ is considerably smaller than usual, since $\epsilon$ (i.e., the infinitesimal step between $x_{near}$ and $x_{new}$ in classic RRT) can be ignored for holonomic problems. However, by adding intermediate nodes, the complexity should rise. In this case, space complexity can be expressed as $\mathcal{O}(n')$ with $n'$, the number of nodes defined in (2) as

$$n' = \sum_{i}^{n} ceil(\frac{\text{dist}(q_{p,i}, q_{c,i})}{\delta}) \simeq n\frac{\overline{\text{dist}}(q_p, q_c)}{\delta}, \quad (2)$$

where $q_{p,i}$ and $q_{c,i}$ are the parent and child nodes of branch $i$, respectively, and $\delta$ is the step size. If we were to choose $\delta = \epsilon$ with a branch distance equal to $\epsilon$ as in classic RRT, $n' = n$ and the space complexity would be the same as RRT. Since the operation of intermediate nodes insertion is bypassed when $\delta = \epsilon$, the time complexity is also unchanged. We can assume that a tree with intermediate nodes on straight line branches is roughly the same as a classic tree with the same infinitesimal step size. The main difference is that it requires fewer random sample generations (none for intermediate nodes) so that $n$ is considerably smaller, as anticipated. However, a bigger number of failed iterations could lead to a rise in $n$, especially in complex problems, leading to a value of $n$ closer to the usual one. Finally, with RRT-Rope starting with the first path found by RRT instead of an optimized path, $n$ is once again considerably reduced.

*2) Path optimization:* Several cases will be used to assess complexity with upper and lower bounds: (a) no shortcut is found; (b) only significantly small shortcuts are found between nodes that are one waypoint apart, resulting in a constant number of waypoints before and after optimization; and (c) a straight-line shortcut exists between $q_I$ and $q_G$.

The time complexity upper bound is $\mathcal{O}(\frac{m\,(m+1)}{2})$, expressed as a number of path waypoints $m$:

$$m = ceil(\frac{c(\tau_f)}{\delta}). \quad (3)$$

This upper bound considers case (a) and (b). These cases are equivalent, since collision checks must be performed by ray casting at a step size that is similar to $\delta$. The operations of node adding and collision checking are very similar in terms of computation time. So, for either case, $1+2+3+...+m$ operations need to be performed to check each pair $(q_i, q_j)$ and the intermediate nodes between them. The lower bound for time complexity is $\mathcal{O}(m_{rope})$ for case (c), with $m_{rope}$ defined as follows:

$$m_{rope} = ceil(\frac{c(\tau_{rope})}{\delta}) \leq m. \quad (4)$$

The upper bound for space complexity is $\mathcal{O}(\frac{m\,(m+1)}{2})$ for case (b), and the lower bound is $\mathcal{O}(0)$ for case (a).

Since the ranges between lower and upper bounds are wide and the value of $m$ depends on $\delta$, the processing time will be analyzed based on simulation results in section V-C.

### *B. Completeness and optimality*

Since probabilistic completeness is inherited from the algorithm used for the search operation, it is ensured with RRT-like algorithms. Thus, one can be sure to find a path if it exists. Also, monotonous convergence is ensured with line 8 of Algorithm 1 and with triangular inequality.

Optimality is an extension of the feasible path planning problem that was defined in section II-C. Let $\Sigma$ be the set of all nontrivial paths. Optimal path planning can be defined similarly to [5] and [6], as the search for a path $\tau^*$ that minimizes the cost function $c : \Sigma_{C_{free}} \rightarrow \mathbb{R}_{\geq 0}$, which assigns a non-negative cost to all nontrivial paths in which each point is on the closure of $C_{free}$. Optimality is ensured when a feasible path with minimal cost is found. Given $C$, $C_{obs}$, $q_I$ and $q_G$, an optimal algorithm must find a path $\tau^* : [0,1] \rightarrow cl(C_{free})$ such that (a) $\tau^*(0) = q_I$ and $\tau^*(1) = q_G$ and (b) $c(\tau^*) = min_{\tau \in \Sigma_{cl(C_{free})}} c(\tau)$, and report failure if no such path exists.

Asymptotic optimality is not guaranteed by RRT-Rope if used as it is, because of the local minimum that can occur in some specific situation. However, the main goal of the algorithm is to speed up the search for a better path. Also, the extension discussed in section VI could ensure asymptotic optimality at the cost of extra computing time.

## V. SIMULATION

The approach was integrated in ROS (Robot Operating System) and tested on simulated environments in Gazebo that were reconstructed from real LiDAR scans. The simulated UAV is equipped with a rotating Ouster LiDAR. A simultaneous localization and mapping (SLAM) approach called RTAB-Map [22] is used to provide the OctoMap and localization inputs for the RRT-Rope algorithm. Simulations were performed with an HP Z440 Workstation with 12 Intel Xeon processors running at 3.5 GHz, 24 GB of RAM, with a loop frequency of 1 MHz for path planning iterations. The simulated environments and associated query pairs $(q_I, q_G)$ are pictured in Fig. 5.

The goal was to reproduce a classic return-to-home operation after UAV exploration. These three environments have been chosen to best represent the spatial characteristics of some typical missions. *Env. 2* represents a classic stope environment, with a cylindrical shape and tunnels at the top. The path should be between 20 and 30 m long. *Env. 1* is a bit different, since the take-off point is farther from the stope and the UAV has to move through a long tunnel first. The stope is also different in that it is more like a long inclined flat space and the full path length is between 30 and 40 m. Finally, *Env. 3* is used to establish how well the algorithm works even in an environment that does not meet all the assumptions stated in section II-A, since it is an indoor environment with columns and doors, with path lengths between 40 and 50 m.

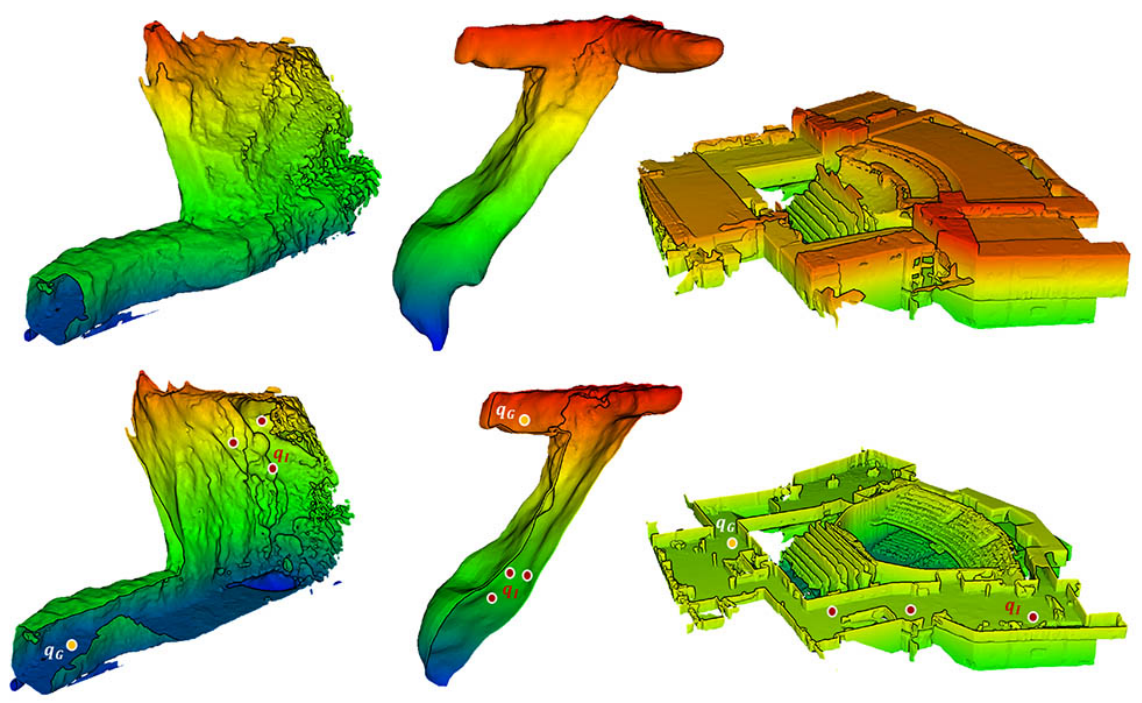

Fig. 5: External (top row) and cross-section (bottom row) views of the three testing environments: 1 (left), 2 (center), and 3 (right). Note that the points $q_I$ and $q_G$ used in the next simulations are respectively in burgundy and yellow.

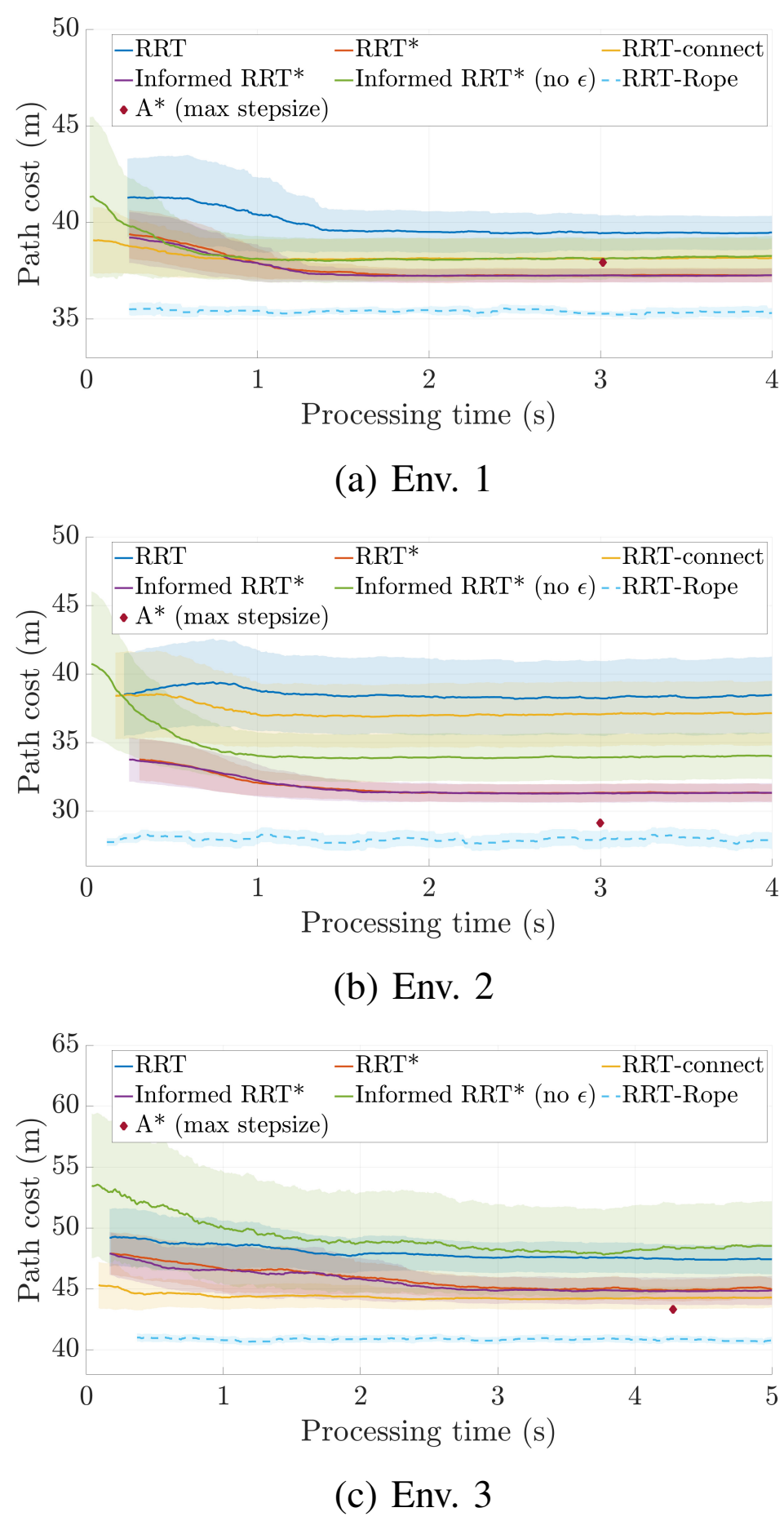

(a) Env. 1

(b) Env. 2

(c) Env. 3

Fig. 6: Moving average of 200 samples of RRT-Rope with $\delta = 0.6\ m$ (cyan), RRT [3] (blue), RRT* [5] (orange), RRT-connect [12] (yellow), Informed RRT* [6] (purple), Informed RRT* without $\epsilon$ (green), and A* [2] (burgundy). The moving average of the standard deviation is shown in shade.

## A. Convergence

First, Fig. 6 shows the results from a simulation used to demonstrate the algorithm's fast convergence. These are the results obtained from running RRT and A* algorithms compared to RRT-Rope optimization applied on a path obtained with Informed RRT* with intermediate nodes and initialized at a different time between the moment a path was found and 5 s. This test had two goals: (a) to confirm the performance of using RRT-Rope optimization on the first feasible path instead of waiting for a better path, and (b) to get a first idea of how a shortened path compares with some path planning algorithms for the same processing time.

The first conclusion can be easily deducted from Figs. 6a, 6b, and 6c, with the RRT-Rope path cost being almost constant with respect to processing time and with a low standard deviation. Since the path cost does not depend on the initial path quality, partially due to the homotopy of paths in our environments, online optimization is obsolete. It shows good confidence in the fact that shortening should be performed after the first feasible path is found, to improve computation time without deteriorating final path cost.

Note that RRT-connect denotes the classic implementation here, while its implementation in RRT-Rope is slightly different. The effect of this variation is noteworthy between the first path finding time of the two Informed RRT* curves.

The path lengths confirm that RRT-Rope outperforms the other planning algorithms for a comparable processing time (i.e., the second objective). A* does not lead to the optimal path because of node resolution that was chosen as low as possible to quickly find the solution. More detailed surveys need to be performed to better understand when RRT-Rope outperforms path shortening algorithms.

## B. Performance comparison

Fig. 7 shows the simulations carried out for 100 samples of each shortening algorithm considered. Unless explicitly stated in the legend, the shortening algorithms were performed on classic RRT-connect paths to find a first path in a fair amount of time.

The partial shortcut with RRT-connect without $\epsilon$ finds a first path more quickly but reaches a plateau far from the optimum because of the low node resolution. Random node pruning also reaches a plateau because of the node resolution which has become low after some nodes were pruned. As expected, elastic strips take much time to converge. Partial shortcut takes more time than RRT-Rope to converge because of irrelevant shortcuts and unnecessarily large node

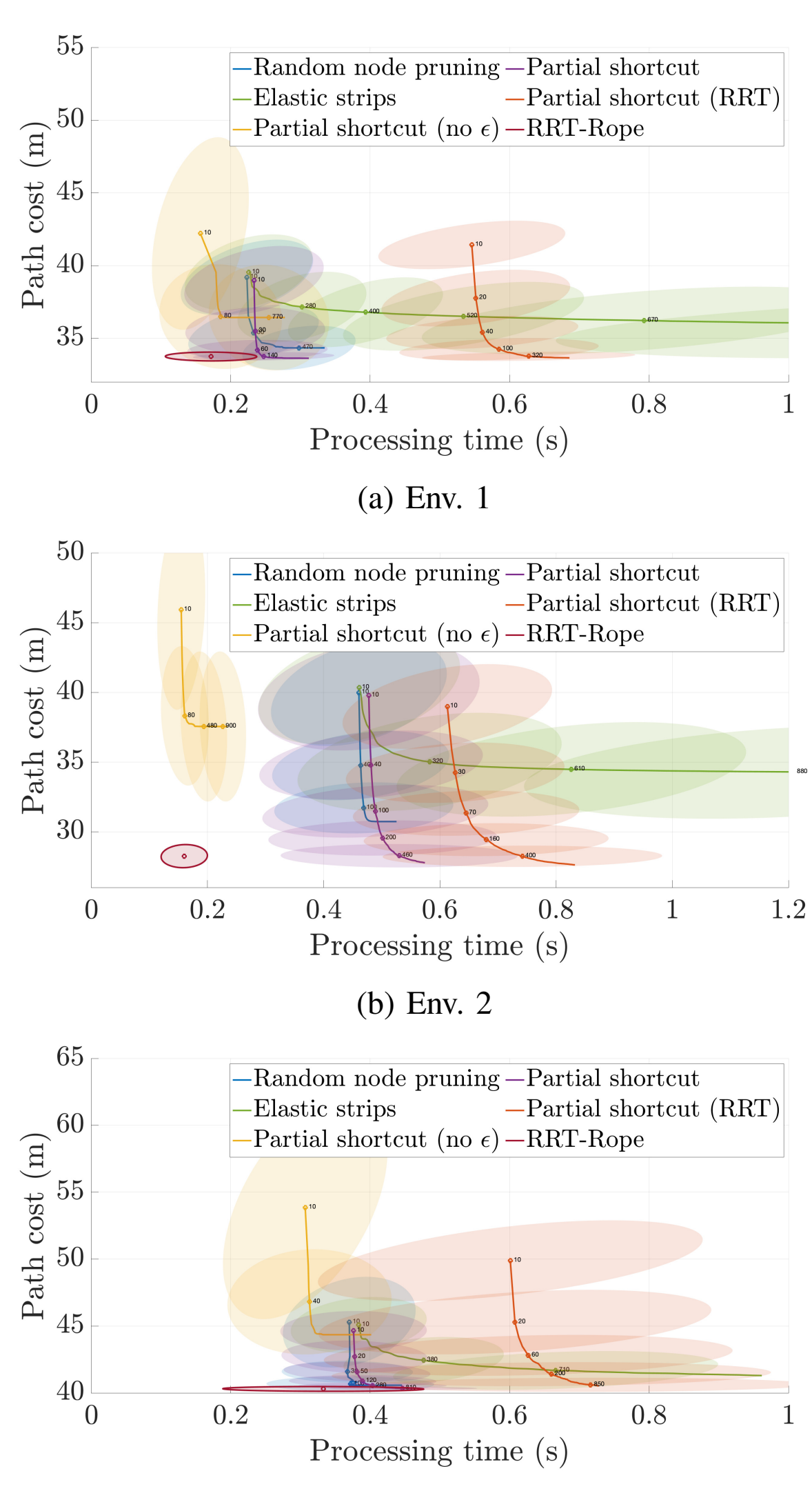

(a) Env. 1

(b) Env. 2

(c) Env. 3

Fig. 7: Standard error ellipse plot for 100 samples of path shortening algorithms of RRT-Rope (burgundy), Random node pruning [14] (blue), Elastic strips [17] (green), Partial shortcut without $\epsilon$ (yellow), Partial shortcut [15] (purple), and Partial shortcut on RRT (red). The number of iterations is indicated next to the corresponding points.

resolution as stated in section I.

In Env. 2—that represent the classic stope environment—RRT-Rope error ellipse presents a big gap with other algorithms, and is 70% faster than the next best algorithm for the same path cost. This gap is reduced in Env. 1 that is narrower. In Env. 3—that was used to challenge the algorithm—the time spread is bigger because of the path finding that is less straightforward. In the end, RRT-Rope shows the best performance in each environment, with a path cost around 1.8% of optimum on average. It also offers the advantage of being a single point, giving the user the information of when the process is finished. In contrast, partial shortcut, i.e. the best second option, reaches its convergence after 150 iterations and 0.25 s in Env. 1, 500 iterations and 0.55 s in Env. 2, and 800 iterations and 0.45 s in Env. 3. It can be a downside to set the maximum number of iterations or processing time considering that convergence is difficult to evaluate online for an a priori unknown path.

### *C. Step-size sensitivity analysis*

Parameter $\delta$ can be adjusted in RRT-Rope. This section evaluates the algorithm's sensitivity to this parameter by reporting results of path cost and processing time, in the same three environments, for values of $\delta$ between 0.05 and 2.05 m. Intuitively, an infinitesimally small value should give the path a perfect rope-like appearance, whereas a larger $\delta$ will transform the trajectory into a chain of longer links.

As shown in (2), the number of nodes in the tree follows approximately the ratio of the sum of the branch lengths over $\delta$. The number of final tree nodes $n_{rope}$ after RRT-Rope can be expressed as

$$n' \leq n_{rope} \leq n' + \frac{m\,(m+1)}{2}, \tag{5}$$

where $n'$ is the number of tree nodes before optimization, and $\frac{m\,(m+1)}{2}$ is the upper bound of the path waypoints added during optimization, with $m$ being the number of waypoints of $\tau_f$ previously defined in (3). $c(\tau_f)$, $n$, and the branch lengths can be regarded as independent of the path optimization step, since they only depend on the search algorithm. By substituting (2) and (3) into (5) and considering that $\delta$ is the only variable, (6) is obtained as follows:

$$\frac{K_1}{\delta} \leq n_{rope} \leq \frac{K_2}{\delta} + \frac{K_3}{\delta^2}. \tag{6}$$

Fig. 8 shows $n_{rope}$ in red, and the number of final path waypoints $m_{rope}$ (defined in (4)) in blue. The red shaded area represents the standard deviation of $n_{rope}$, which is caused by (a) randomness of $c(\tau_f)$, $n$, and the branch lengths, and (b) variations inside the limit bounds of (6). The nearly nonexistent blue shaded area is due to variations of $c(\tau_{rope})$, which are shown to be quite low in Fig. 6 for a given $\delta$.

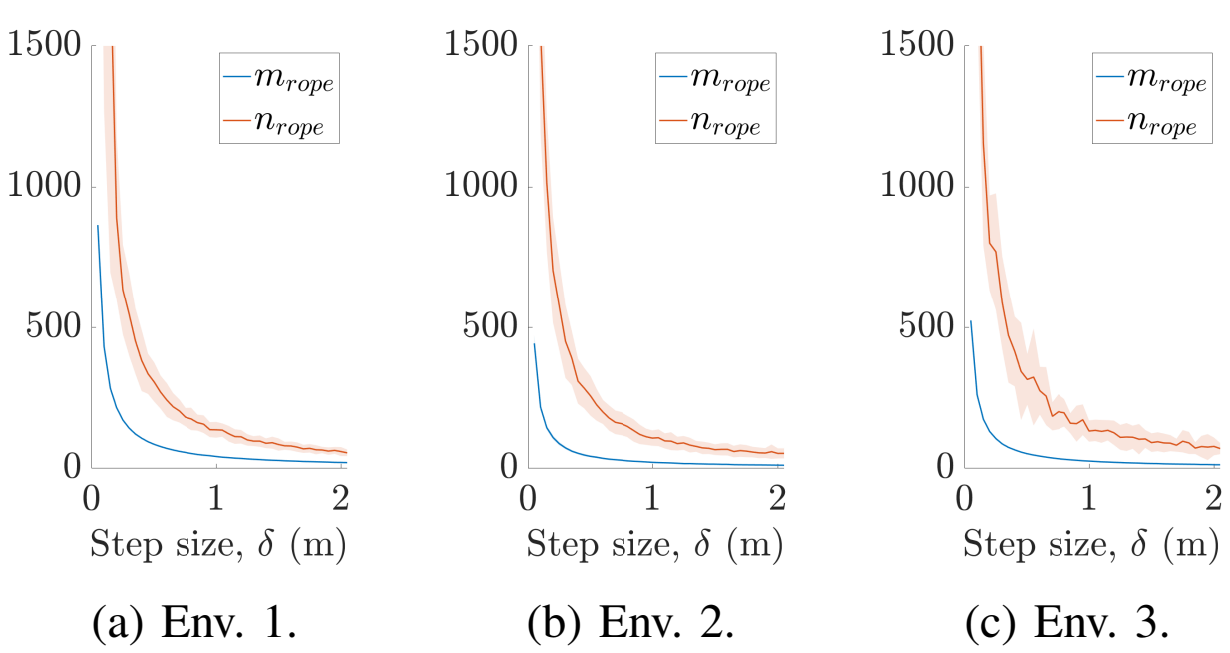

(a) Env. 1. (b) Env. 2. (c) Env. 3.

Fig. 8: Number of path waypoints (blue) and tree nodes (red) for different $\delta$ values.

Also, variations in Fig. 9 show that for $\Delta\delta < 1.94\,\overline{\delta}$, the path cost variation $\Delta c < 0.08\,\overline{c}$. Therefore, the curve of path waypoints can be considered as following an inverse function of $\delta$ if $\Delta c$ is neglected.

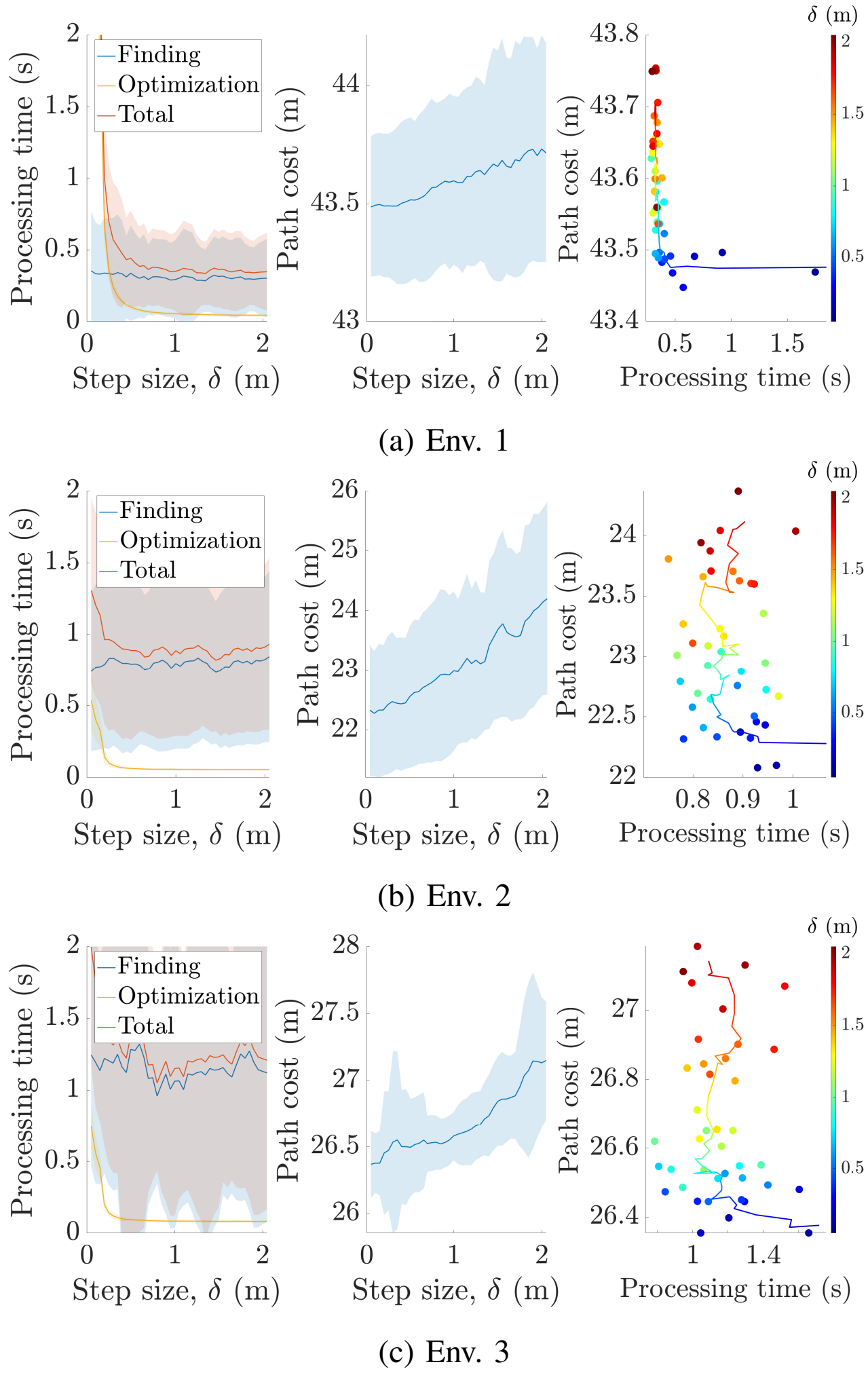

Fig. 9: Processing time (left), path cost (center), and total (right) for different $\delta$ values. The colored curve represents the moving average.

One can imagine that increasing the number of tree nodes by reducing $\delta$ would lead to a longer processing time. Fig. 9 shows the experimental processing time for path optimization in yellow, the total time in red, and the path finding time in blue. The path cost is represented in the center column of the figures. As expected, processing time decreases as a function of $\delta$, whereas path cost grows with it.

The figure on the right side of Fig. 9 shows the same results on a cost-time axis. By looking at the moving average curve, we can see that $\delta$ generally enables us to make a trade-off between processing time and path cost.

To find an optimal $\delta$ minimizing the total time to move from $q_I$ to $q_G$, a cost function of total flight time (including processing and travel) is introduced as

$$g_t = \frac{c(\tau)}{v} + t, \tag{7}$$

where $v$ is the UAV velocity, $t$ is the processing time, and $c(\tau)$ is the path cost. Fig. 10 shows the standardized value $g$ of the cost function $g_t$ for the sake of readability. Indeed, the flight duration is longer for slower UAVs, but one cares only about finding the value of $\delta$ minimizing the cost function. The minimum points of each curve show that a smaller value of $\delta$ is better for slow UAVs, whereas for faster UAVs it is preferable to have a lower computation time at the expense of a higher path cost. It is intuitively clear that a fast vehicle has an interest in starting its movement quickly even if it has a slightly longer path to cover, whereas a slow vehicle has an interest in spending more time optimizing its route.

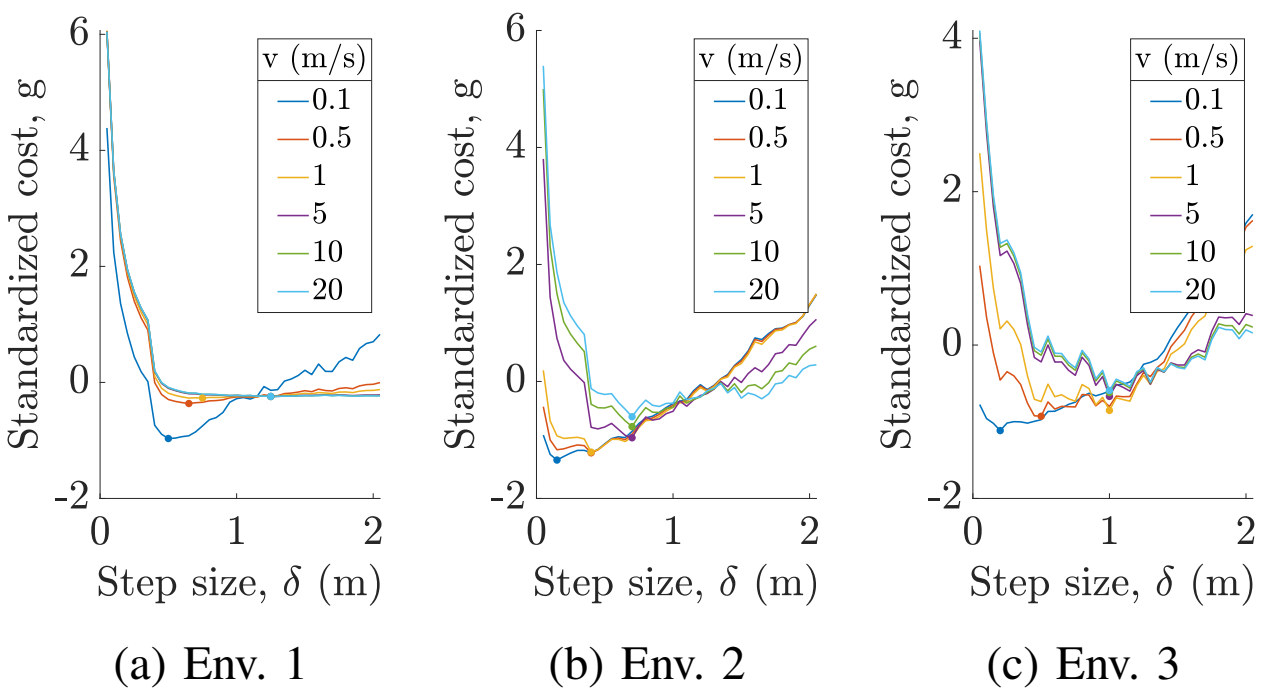

Fig. 10: Standardized cost $g$ as a function of $\delta$ for different UAV velocities.

The minimum points $\delta^*$ are reported in Fig. 11. It can be observed that one should adopt a larger value of $\delta$ when the distance between $q_I$ and $q_G$ is higher. We also observed qualitatively that one should increase $\delta$ for a larger average cross-sectional area of the environment. However, this metric is more difficult to compute. Also, $g$ has a relatively flat curve for each figure in the area where $\delta$ is between 0.5 and 1.5 m. This may motivate the user to set a constant value around 0.8 m for good performance in many environments.

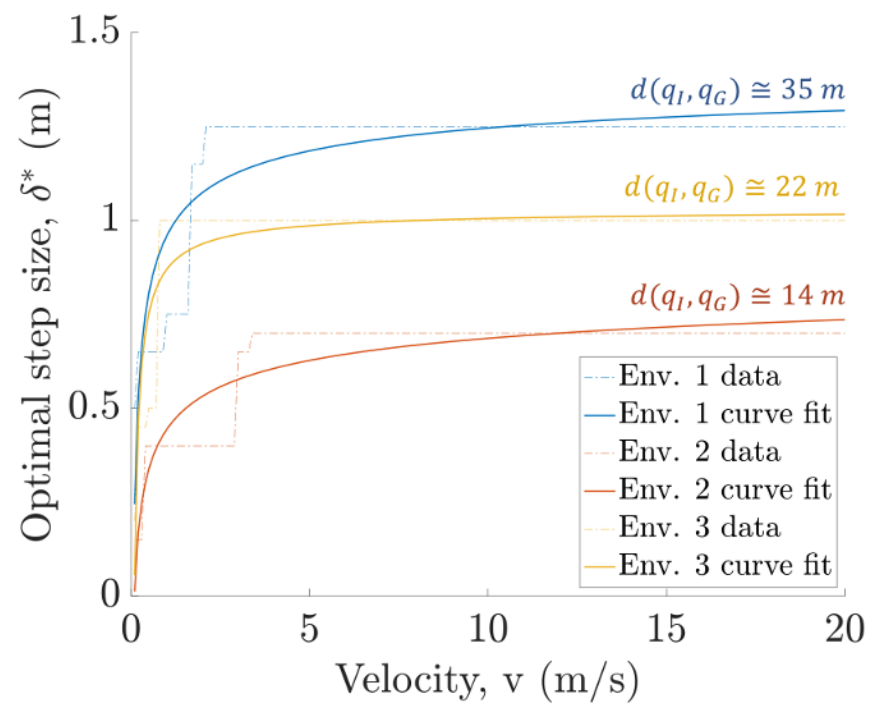

Fig. 11: Value of $\delta^*$ optimizing standardized cost as a function of UAV velocities for different environments.

## VI. DISCUSSION

Interestingly, pillar-like structures would not present a problem for RRT-Rope even if they would for a physical rope, as illustrated in Fig. 11 (right). In the same way, the algorithm could find a new homotopy in an uncluttered environment with wide passages. However, RRT-Rope could face local minima problems in complex environments, as illustrated in Fig. 12 (left). In mine stopes, this problem does not arise because the walls are smooth from the macro point of view.

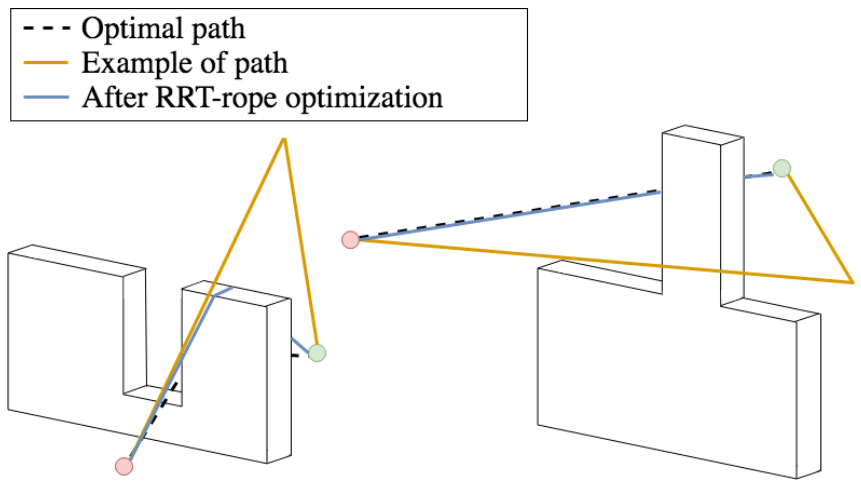

Fig. 12: Representation of local minima situations (left) and non-problematic situations (right) with RRT-Rope.

The main challenge in resolving this hypothetical issue is detecting the local minima. A potential strategy for doing so is to use Informed-RRT* to escape local minima and to estimate (a) if the configuration has escaped a local minimum by monitoring the path cost decrease, and (b) if the algorithm has not yet converged thanks to the path cost decrease rate.

The other approach to improving RRT-Rope is to include dynamic constraints in the problem definition. This could be done by detecting turns and imposing angle restriction constraints similarly to [23], or by using a local planner coupled with RRT-Rope.

## VII. Conclusion

The presented RRT-Rope algorithm can solve near-optimal path planning problems in a short amount of time for large environments. The algorithm's speed is inherited from an altered RRT-connect algorithm's fast time to compute a feasible path, coupled to a fast rope-inspired optimization algorithm for a random trajectory. Path finding time is improved by deleting infinitesimal step size which reduces random node generations needed in large uncluttered environments. The first path is sufficient for rope shortening algorithm which performance has been shown to be almost independent from the input path quality in 3-ball homotopic environments. Irrelevant shortcuts are avoided by checking farthest nodes first and detecting straight lines. Intermediate nodes insertion ensures near-optimality and uniform resolution. RRT-Rope produces an equal or shorter path than the other presented algorithms in a shorter computation time, and consistent behaviours without the need for a varying number of iterations thanks to deterministic selection.

The results are loosely sensitive to the environment and to the step size $\delta$. This latter parameter can be optimized for a given path cost and robot velocity. Future works include introducing dynamic constraints and handling more complex environments.

## Acknowledgment

The authors would like to thank Dr. Mathieu Labbé (IntRoLab, University of Sherbrooke) for his help with implementing the algorithm in the ROS system.

## References

[1] L. Yang, J. Qi, J. Xiao, and X. Yong, "A literature review of UAV 3D path planning," *Proceeding of the 11th World Congress on Intelligent Control and Automation*, pp. 2376–2381, 2014.

[2] P. E. Hart, N. J. Nilsson, and B. Raphael, "A formal basis for the heuristic determination of minimum cost paths," *IEEE Transactions on Systems Science and Cybernetics*, vol. 4, no. 2, pp. 100–107, 1968.

[3] S. R. Lindemann and S. M. LaValle, "Randomized Kinodynamic Planning," *The International Journal of Robotics Research*, vol. 20, no. 5, pp. 378–400, 2001.

[4] S. M. LaValle and J. Kuffner, "Rapidly-Exploring Random Trees: Progress and Prospects," *In Algorithmic and Computational Robotics: New Directions*, pp. 293–308, 2000.

[5] S. Karaman and E. Frazzoli, "Incremental Sampling-Based Algorithms for Optimal Motion Planning," *The International Journal of Robotics Research*, vol. 30, no. 7, pp. 846–894, 2010.

[6] J. D. Gammell, S. S. Srinivasa, and T. D. Barfoot, "Informed RRT*: Optimal sampling-based path planning focused via direct sampling of an admissible ellipsoidal heuristic," *IEEE/RSJ International Conference on Intelligent Robots and Systems*, pp. 2997–3004, 2014.

[7] O. Arslan and P. Tsiotras, "Use of relaxation methods in sampling-based algorithms for optimal motion planning," *IEEE International Conference on Robotics and Automation*, pp. 2421–2428, 2013.

[8] M. Otte and N. Correll, "C-FOREST: Parallel Shortest Path Planning With Superlinear Speedup," *IEEE Transactions on Robotics*, vol. 29, no. 3, pp. 798–806, 2013.

[9] M. Kanehara, S. Kagami, J. J. Kuffner, S. Thompson, and H. Mizoguhi, "Path shortening and smoothing of grid-based path planning with consideration of obstacles," *IEEE International Conference on Systems, Man and Cybernetics*, pp. 991–996, 2007.

[10] S. Karaman, M. R. Walter, A. Perez, E. Frazzoli, and S. Teller, "Anytime Motion Planning using the RRT*," *IEEE International Conference on Robotics and Automation*, pp. 1478–1483, 2011.

[11] J. D. Gammell, T. D. Barfoot, and S. S. Srinivasa, "Informed Sampling for Asymptotically Optimal Path Planning," *IEEE Transactions on Robotics*, vol. 34, no. 4, pp. 966–984, 2018.

[12] J. Kuffner and S. M. LaValle, "RRT-Connect: An Efficient Approach to Single-Query Path Planning," *IEEE International Conference on Robotics and Automation*, vol. 2, pp. 995–1001, 2000.

[13] J. Luo and K. Hauser, "An empirical study of optimal motion planning," *IEEE/RSJ International Conference on Intelligent Robots and Systems*, pp. 1761–1768, 2014.

[14] F. Islam, J. Nasir, U. Malik, Y. Ayaz, and O. Hasan, "RRT*-Smart: Rapid convergence implementation of RRT* towards optimal solution," *IEEE International Conference on Mechatronics and Automation*, pp. 1651–1656, 2012.

[15] R. Geraerts and M. H. Overmars, "Clearance based path optimization for motion planning," *IEEE International Conference on Robotics and Automation, 2004*, vol. 3, pp. 2386–2392, 2004.

[16] S. Berchtold and B. Glavina, "A scalable optimizer for automatically generated manipulator motions," *IEEE/RSJ International Conference on Intelligent Robots and Systems*, vol. 3, pp. 1796–1802, 1994.

[17] O. Brock and O. Khatib, "Elastic strips: A framework for motion generation in human environments," *The International Journal of Robotics Research*, vol. 21, no. 12, pp. 1031–1052, 2002.

[18] S. M. LaValle, *Planning Algorithms*. Cambridge University Press, 2006.

[19] A. Hornung, K. M. Wurm, M. Bennewitz, C. Stachniss, and W. Burgard, "OctoMap: An Efficient Probabilistic 3D Mapping Framework Based on Octrees," *Autonomous Robots*, 2013.

[20] R. Luna, I. A. Şucan, M. Moll, and L. E. Kavraki, "Anytime solution optimization for sampling-based motion planning," *IEEE International Conference on Robotics and Automation*, pp. 5068–5074, 2013.

[21] S. Karaman and E. Frazzoli, "Sampling-based Algorithms for Optimal Motion Planning," *International Journal of Robotic Research*, vol. 30, pp. 846–894, 2011.

[22] M. Labbé and F. Michaud, "RTAB-Map as an open-source lidar and visual simultaneous localization and mapping library for large-scale and long-term online operation," *Journal of Field Robotics*, vol. 36, 2018.

[23] J. Tian, Y. Wang, and D. Yuan, "An Unmanned Aerial Vehicle Path Planning Method Based on the Elastic Rope Algorithm," *IEEE 10th International Conference on Mechanical and Aerospace Engineering*, pp. 137–141, 2019.